# A Qualitative Markov Assumption and Its Implications for Belief Change

**Nir Friedman**
Stanford University
Dept. of Computer Science
Stanford, CA 94305-9010
nir@cs.stanford.edu

**Joseph Y. Halpern**
IBM Almaden Research Center
650 Harry Road
San Jose, CA 95120-6099
halpern@almaden.ibm.com

## Abstract

The study of *belief change* has been an active area in philosophy and AI. In recent years, two special cases of belief change, *belief revision* and *belief update*, have been studied in detail. Roughly speaking, revision treats a surprising observation as a sign that previous beliefs were wrong, while update treats a surprising observation as an indication that the world has changed. In general, we would expect that an agent making an observation may both want to revise some earlier beliefs and assume that some change has occurred in the world. We define a novel approach to belief change that allows us to do this, by applying ideas from probability theory in a qualitative settings. The key idea is to use a qualitative Markov assumption, which says that state transitions are independent. We show that a recent approach to modeling qualitative uncertainty using *plausibility measures* allows us to make such a qualitative Markov assumption in a relatively straightforward way, and show how the Markov assumption can be used to provide an attractive belief-change model.

## 1 INTRODUCTION

The question of how an agent should change his beliefs after making an observation or performing an action has attracted a great deal of recent attention. There are two proposals that have received perhaps the most attention: *belief revision* [Alchourrón, Gärdenfors, and Makinson 1985; Gärdenfors 1988] and *belief update* [Katsuno and Mendelzon 1991]. Belief revision focuses on how an agent changes his beliefs when he acquires new information; *belief update*, on the other hand, focuses on how an agent should change his beliefs when he realizes that the world has changed. Both approaches attempt to capture the intuition that to accommodate a new belief, an agent should make minimal changes. The difference between the two approaches comes out most clearly when we consider what happens when an agent observes something that is inconsistent with his previous beliefs. Revision treats the new observation as an indication that some of the previous beliefs were wrong and should be discarded. It tries to choose the most plausible beliefs that can accommodate the observation. Update, on the other hand, assumes that the previous beliefs were correct, and that the observation is an indication that a change occurred in the world. It then tries to find the most plausible change that accounts for the observation and to predict what else has changed as a result.

In general, we would expect that an agent making an observation may want both to revise some earlier beliefs and to assume that some change has occurred in the world. To see this, consider the following example (which is a variant of Kautz's *stolen car* example [1986], and closely resembles the *borrowed-car* example in [Friedman and Halpern 1994b]): A car is parked with a full fuel tank at time 0; at time 2, the owner returns to find it parked where he left it. If the owner believes that parked cars tend to stay put, then he would believe that no changes occurred between time 0 and 2. What should he believe when, at time 3, he discovers that the fuel tank is empty? Update treats this observation as an indication of a change between time 2 and 3, for example, a gas leak. Revision, on the other hand, treats it as an indication that previous beliefs, such as the belief that the tank was full at time 2, were wrong. In practice, the agent may want to consider a number of possible explanations for his time-3 observation, depending on what he considers to be the most likely sequence(s) of events between time 0 and time 3. For example, if he has had previous gas leaks, then he may consider a gas leak to be the most plausible explanation. On the other hand, if his wife also has the car keys, he may consider it possible that she used the car in his absence. Is there a reasonable approach that lets us capture these considerations in a natural manner? In this paper, we show that there is and, indeed, we can get one by applying straightforward ideas from probability theory in a qualitative setting.

To understand our approach, it is helpful to review what a probabilist would do. The first step is to get an appropriate model of the situation. As was argued in [Friedman and Halpern 1995a; Friedman and Halpern 1994b], to capture belief change appropriately, we need a model of how the system changes over time. We assume that at each point in time, the system is in some *state*. A *run* of the system is a function from time (which we assume ranges over the natural numbers) to states; thus, a run is essentially a sequence of states. A run can be thought of as a description of how the system changes over time. We identify a system with a set of runs. Intuitively, we are identifying the system with its possible behaviors.

The standard probabilistic approach would be to put a probability on the runs of the system. This is the agent's prior probability, and captures his initial beliefs about the relative likelihood of runs. As the agent receives information, he changes his beliefs using conditioning.

One obvious problem with this approach is that, even if there are only two possible states, there are uncountably many possible runs. How can an agent describe a prior probability over such a complex space? The standard solution to this problem is to assume that state transitions are independent of when they occur, that is, that the probability of the system going from state $s$ to state $s'$ is independent of the sequence of transitions that brought the system to state $s$. This *Markov assumption* significantly reduces the complexity of the problem. All that is necessary is to describe the probability of state transitions. Moreover, the Markov assumption has been shown to be widely applicable in practice [Kemeny and Snell 1960; Howard 1971].

Another problem with a straightforward use of probability is that, in many situations, an agent may not know the exact probability of various state transitions, although he may have some more qualitative information about them. In the literature, there are many approaches to representing qualitative beliefs: preferential structures [Kraus, Lehmann, and Magidor 1990], possibilistic measures [Dubois and Prade 1990], $\kappa$-rankings [Spohn 1988; Goldszmidt and Pearl 1992], and logics of extreme probabilities [Pearl 1989]. We represent beliefs here using *plausibility measures* [Friedman and Halpern 1995b], an approach that generalizes all the earlier approaches. A plausibility measure is a qualitative analogue of a probability measure; it associates with every event its *plausibility*, which is just an element in a partially ordered space.

As shown in [Friedman and Halpern 1995a], we can define a natural notion of belief using plausibility, where a proposition is believed exactly if it is more plausible than its complement. It is also easy to define a notion of conditioning in plausibility spaces (as done in [Friedman and Halpern 1995b]). Once we apply conditioning to the notion of belief, we get a notion of belief change. Interestingly, it can be shown that belief revision and belief update both can be viewed as instances of such belief change [Friedman and Halpern 1994b]. That is, we can get belief revision and belief update when we condition on the appropriate plausibility measures. Not surprisingly, the plausibility measures that capture belief revision are ones that consider plausible only runs where the world never changes over time. On the other hand, the plausibility measures that capture belief update are ones that make plausible those runs in which, in a precise sense, abnormalities are deferred as late as possible.

The plausibility measures that give us belief revision and belief update are fairly special, and do not capture many typical situations. We would like to specify a prior plausibility measure over runs that captures our initial assessment of the relative plausibility of runs. As in the probabilistic settings, such a prior can be quite complex. We can use (a qualitative analogue of) the Markov assumption to simplify the description of the prior plausibility.

Making a (qualitative) Markov assumption gives us a well-behaved notion of belief change, without making the occasionally unreasonable assumptions made by belief revision and update. In particular, it allows a user to weigh the relative plausibility that a given observation is due to a change in the world or due to an inaccuracy in previous beliefs. In the car example, this means that the agent can decide the relative plausibility of a gas leak and his wife's taking the car, without making the commitment to one or the other, as required by update and revision.

This paper is organized as follows. In Section 2, we review the probabilistic approach. Then, in Section 3, we review the definition of plausibility and discuss conditional plausibility. In Section 4, we introduce Markovian plausibility measures and show how they can be used to induce an interesting notion of belief change. In Section 5, we examine the situation where the user is willing only to compare the plausibility of transitions, without committing to their magnitude. We characterize what beliefs follow from such a partial specification. In Section 6, we compare our approach to others in the literature. We end with a discussion of these results and directions of future research in Section 7.

## 2   PROBABILISTIC BELIEF CHANGE

To reason about a space $W$ probabilistically, we need a probability measure on $W$. Formally, that means we have a probability space $(W, \mathcal{F}, \Pr)$, where $\mathcal{F}$ is an *algebra* of *measurable* subsets of $W$ (that is, a set of subsets closed under union and complementation, to which we assign probability) function mapping each *event* (i.e., a subset of $W$) in $\mathcal{F}$ to a number in $[0, 1]$, satisfying the well-known Kolmogorov axioms ($\Pr(\emptyset) = 0$, $\Pr(W) = 1$, and $\Pr(A \cup B) = \Pr(A) + \Pr(B)$ if $A$ and $B$ are disjoint).[1]

Probability theory also dictates how we should change our beliefs. If the agent acquires evidence $E$, his beliefs after learning $E$ are specified by the *conditional probability* $\Pr(\cdot|E)$. Note that by using conditioning, we are implicitly assuming that the information $E$ is correct (since we assign $\overline{E}$—the complement of $E$—probability 0), and that discovering $E$ is telling us only that $\overline{E}$ is impossible; the relative probability of subsets of $E$ is the same before and after conditioning.

We want to reason about a dynamically changing system. To do so, we need to identify the appropriate space $W$ and the events of interest (i.e., $\mathcal{F}$). We assume that the system changes in discrete steps, and that after each step, the system is in some *state*. We denoted by $S$ the set of possible states of the system. As we said in the introduction, a run is a function from the natural numbers to states. Thus, a run $r$ describes a particular execution that goes through the sequence of states $r(0), r(1), \ldots$. We identify a system with a set of runs, and take $W$ to consist of these runs.

---
[1] For ease of exposition, we do not make the requirement that probability distributions be countably additive here.

There are various events that will be of interest to us; we always assume without comment that the events of interest are in $\mathcal{F}$. One type of event of interest is denoted $S_i = s$; this is the set of runs $r$ such that $r(i) = s$.[2] A *time-n* event is a Boolean combination of events of the form $S_i = s$, for $i \leq n$. We are particularly interested in time-$n$ events of the form $(S_0 = s_0) \cap \cdots \cap (S_n = s_n)$, which we abbreviate $[s_0, \ldots, s_n]$; this is the set of all runs in $W$ with initial prefix $s_0, \ldots, s_n$. We call such an event an *n-prefix*. Note that any time-$n$ event is a union of $n$-prefixes.

As discussed in the introduction, describing a distribution Pr on runs can be difficult. Even when $\mathcal{S}$ contains only two states, $W$ is uncountable. In the probabilistic literature, this difficulty is often dealt with by making a *Markov assumption*.

**Definition 2.1:** A *Markov chain* [Kemeny and Snell 1960] over $S_1, S_2, \ldots$ is a measure Pr on $W$ that satisfies

- $\Pr(S_{n+1} = s_{n+1} \mid E, S_n = s_n) = \Pr(S_{n+1} = s_{n+1} \mid S_n = s_n)$, where $E$ is any time-$n$ event,
- $\Pr(S_{m+1} = s' \mid S_m = s) = \Pr(S_{n+1} = s' \mid S_n = s)$.

We say that Pr is a *Markovian* measure if it is a Markov chain over $S_0, S_1, \ldots$. ∎

The first requirement states that the probability of $S_{n+1} = s_{n+1}$ is independent of preceding states given the value of $S_n$: The probability of going from state $S_n = s_n$ to $S_{n+1} = s_{n+1}$ is independent of *how* the system reached $S_n = s_n$. The second requirement is that the *transition probabilities*, i.e., the probabilities of transition from state $s$ to state $s'$, do not depend on the time of the transition. Many systems can be modeled so as to make both assumptions applicable.

If we assume that the system has a unique initial state $s_0$ (that is, $r(0) = s_0$ for all runs $r \in W$), and specify transition probabilities $p_{s,s'}$, with $\sum_{s' \in \mathcal{S}} p_{s,s'} = 1$, for each $s \in \mathcal{S}$, then it is easy to show that there is a unique Markovian measure Pr on the algebra generated by events of the form $S_j = s$ such that $\Pr(S_{n+1} = s' \mid S_n = s) = p_{s,s'}$. We can define Pr on the $n$-prefixes by induction on $n$. The case of $n = 0$ is given by $\Pr([s_0]) = 1$. For the inductive step, assuming that we have defined $\Pr([s_0, \ldots, s_{n-1}])$, we have

$$\begin{aligned}
&\Pr([s_0, \ldots, s_n]) \\
&= \Pr(S_n = s_n \mid [s_0, \ldots, s_{n-1}]) \times \Pr([s_0, \ldots, s_{n-1}]) \\
&= p_{s_{n-1}, s_n} \times \Pr([s_0, \ldots, s_{n-1}]).
\end{aligned}$$

An easy induction argument now shows

$$\Pr([s_0, \ldots, s_n]) = p_{s_0, s_1} \times \cdots \times p_{s_{n-1}, s_n}.$$

---

[2]Technically, $S_i$ is a *random variable* taking on values in $\mathcal{S}$, the set of states.

Since a time-$n$ event is a union of $n$-prefixes, this shows that Pr is determined for all time-$n$ events.

If $\mathcal{S}$ is finite, this gives us a compact representation of the distribution over $W$. Of course, even if we do not have a unique initial state, we can construct a Markovian probability distribution from the transition probabilities and the probabilities $\Pr([s])$ for $s \in \mathcal{S}$. For ease of exposition, throughout this paper, we make the following simplifying assumption.

> **Simplifying assumption:** $\mathcal{S}$ contains the state $s_0$ and $W$ is a set of runs over $\mathcal{S}$, all of which have initial state $s_0$.

Obvious analogues of our results hold even without this assumption.

## 3 PLAUSIBILITY MEASURES

Our aim is to find analogues of probabilistic belief change in situations where we do not have numeric probabilities. We do so by using notion of a *plausibility space*, which is a natural generalization of probability space [Friedman and Halpern 1995b]. We simply replace the probability measure Pr by a *plausibility measure* Pl, which, rather than mapping sets in $\mathcal{F}$ to numbers in $[0, 1]$, maps them to elements in some arbitrary partially ordered set. We read $\text{Pl}(A)$ as "the plausibility of set $A$". If $\text{Pl}(A) \leq \text{Pl}(B)$, then $B$ is at least as plausible as $A$. Formally, a *plausibility space* is a tuple $S = (W, \mathcal{F}, D, \text{Pl})$, where $W$ is a set of worlds, $\mathcal{F}$ is an algebra of subsets of $W$, $D$ is a domain of *plausibility values* partially ordered by a relation $\leq_D$ (so that $\leq_D$ is reflexive, transitive, and anti-symmetric), and Pl maps the sets in $\mathcal{F}$ to $D$. We assume that $D$ is *pointed*: that is, it contains two special elements $\top_D$ and $\bot_D$ such that $\bot_D \leq_D d \leq_D \top_D$ for all $d \in D$; we further assume that $\text{Pl}(W) = \top_D$ and $\text{Pl}(\emptyset) = \bot_D$. The only other assumption we make is

**A1.** If $A \subseteq B$, then $\text{Pl}(A) \leq \text{Pl}(B)$.

Thus, a set must be at least as plausible as any of its subsets. As usual, we define the ordering $<_D$ by taking $d_1 <_D d_2$ if $d_1 \leq_D d_2$ and $d_1 \neq d_2$. We omit the subscript $D$ from $\leq_D$, $<_D$, $\top_D$, and $\bot_D$ whenever it is clear from context.

Clearly plausibility spaces generalize probability spaces. They also are easily seen to generalize Dempster-Shafer *belief functions* [Shafer 1976] and *fuzzy measures* [Wang and Klir 1992], including *possibility measures* [Dubois and Prade 1990]. Of more interest to us here is another approach that they generalize: An *ordinal ranking* (or $\kappa$*-ranking*) on $W$ (as defined by [Goldszmidt and Pearl 1992], based on ideas that go back to [Spohn 1988]) is a function $\kappa : 2^W \to \mathbb{N}^*$, where $\mathbb{N}^* = \mathbb{N} \cup \{\infty\}$, such that $\kappa(W) = 0$, $\kappa(\emptyset) = \infty$, and $\kappa(A) = \min_{a \in A} \kappa(\{a\})$ if $A \neq \emptyset$. Intuitively, an ordinal ranking assigns a degree of surprise to each subset of worlds in $W$, where 0 means unsurprising, and higher numbers denote greater surprise. Again, it is easy to see that if $\kappa$ is a ranking on $W$, then $(W, \mathbb{N}^*, \kappa)$ is a plausibility space, where $x \leq_{\mathbb{N}^*} y$ if and only if $y \leq x$ under the usual ordering on the ordinals.

Conditioning plays a central role in probabilistic belief change. In [Friedman and Halpern 1995b], we define an analogue of conditioning for plausibility. Just as a conditional probability measure associates with each pair of sets $A$ and $B$ a number, $\Pr(A|B)$, a conditional plausibility measure associates with pairs of sets a conditional plausibility. Formally, a conditional plausibility measure maps maps a pair of sets $A$ and $B$ to a plausibility, usually denoted $\text{Pl}(A|B)$, where for each fixed $B \neq \emptyset$, $\text{Pl}(\cdot|B)$ is a plausibility measure, satisfying a coherence condition described below. A *conditional plausibility space* is a tuple $(W, \mathcal{F}, D, \text{Pl})$, where Pl is a conditional plausibility measure. In keeping with standard practice in probability theory, we also write $\text{Pl}(A, B | D, E)$ rather than $\text{Pl}(A \cap B | D \cap E)$. The coherence condition is

**C1.** $\text{Pl}(A|C, E) \leq \text{Pl}(B|C, E)$ if and only if $\text{Pl}(A, C|E) \leq \text{Pl}(B, C|E)$.

C1 captures the relevant aspects of probabilistic conditioning: after conditioning by $C$, the plausibility of sets that are disjoint from $C$ becomes $\bot$, and the relative plausibility of sets that are subsets of $C$ does not change.

As we mentioned in the introduction, we are interested here in plausibility measures that capture certain aspects of qualitative reasoning. We say that an event $A$ is *believed* given evidence $E$ according to plausibility measure Pl if $\text{Pl}(A, E) > \text{Pl}(\overline{A}, E)$, that is, if $A$ is more plausible than its complement when $E$ is true. Notice that, by C1, this is equivalent to saying that $\text{Pl}(A|E) > \text{Pl}(\overline{A}|E)$. Moreover, note that if Pl is a probability function, this just says that $\Pr(A|E)$ is greater than $1/2$. Probabilistic beliefs defined this way are, in general, not closed under conjunction. We may believe $A$ and believe $A'$ without believing $A \cap A'$. In [Friedman and Halpern 1996], we show that a necessary and sufficient condition for an agent's beliefs to be closed under conjunction is that the plausibility measure satisfies the following condition:

**A2.** If $A$, $B$, and $C$ are pairwise disjoint sets, $\text{Pl}(A \cup B) > \text{Pl}(C)$, and $\text{Pl}(A \cup C) > \text{Pl}(B)$, then $\text{Pl}(A) > \text{Pl}(B \cup C)$.

Plausibility measures that satisfy A2 are called *qualitative*.[3] We can now state precisely the property captured by A2. Given a plausibility measure Pl, let $Bel_{\text{Pl}}(E) = \{A : \text{Pl}(A|E) > \text{Pl}(\overline{A}|E)\}$. We then have:

**Theorem 3.1:** [Friedman and Halpern 1996][4] Pl *is a qualitative plausibility measure if and only if, for all events $A$, $B$, and $E$, if $A, B \in Bel_{\text{Pl}}(E)$ then $A \cap B \in Bel_{\text{Pl}}(E)$.*

---

[3] In [Friedman and Halpern 1996], we also assumed that qualitative plausibility measures had an additional property: if $\text{Pl}(A) = \text{Pl}(B) = \bot$, then $\text{Pl}(A \cup B) = \bot$. This property plays no role in our results, so we do not assume it explicitly here. (In fact, it follows from assumptions we make in the next section regarding decomposability.)

[4] This result is a immediate corollary of [Friedman and Halpern 1996, Theorem 5.4]. The same result was discovered independently by Dubois and Prade [1995].

It is easy to show that possibility measures and $\kappa$-rankings define qualitative plausibility spaces. In addition, as we show in [Friedman and Halpern 1996], preferential orderings [Kraus, Lehmann, and Magidor 1990], and PPDs (parameterized probability distributions, which are used in defining $\epsilon$-semantics [Pearl 1989]) can be embedded into qualitative plausibility spaces. On the other hand, probability measures and Dempster-Shafer belief functions are in general not qualitative. Since our interest here is in qualitative reasoning, we focus on qualitative plausibility spaces (although some of our constructions hold for arbitrary plausibility measures).

Using qualitative (conditional) plausibility spaces we can model belief change in dynamic systems. Both revision and update are concerned with beliefs about the current state of the world. We follow this tradition, although most of our results also apply to richer languages (which allow, for example, beliefs about past and future states). Suppose $(W, \mathcal{F}, D, \text{Pl})$ is a plausibility space, where $W$ is a system consisting of all the runs over some state space $\mathcal{S}$. Informally, we want to think of a language that makes statements about states. That means that each formula in such a language can be identified with a set of states. In particular, a proposition (set of states) $A$ in such a language is true at time $n$ in a run $r$ if $r(n) \in A$. Using the notion of belief defined earlier, $A$ is believed to be true at time $n$ if the plausibility of the set of runs where $A$ is true at time $n$ is greater than the plausibility of the set of runs where it is false. To make this precise, if $A$ is a set of states, let $A^{(n)} = \{r \in W : r(n) \in A\}$. Thus, $A^{(n)}$ is the set of runs where $A$ is true at time $n$. Then we define

$$Bel^n_{\text{Pl}}(E) =_{\text{def}} \{A \subseteq \mathcal{S} \mid \text{Pl}(A^{(n)}|E) > \text{Pl}(\overline{A}^{(n)}|E)\}.$$

We can think of $Bel^n_{\text{Pl}}(E)$ as characterizing the agent's beliefs about the state of the world at time $n$, given evidence $E$. (We omit the subscript Pl from $Bel^n_{\text{Pl}}$ whenever it is clear from the context.)

This construction—which essentially starts with a prior on runs and updates it by conditioning in the light of new information—is analogous to the probabilistic approach for handling observations.

We can also relate our approach to the more standard approaches to belief change [Alchourrón, Gärdenfors, and Makinson 1985; Katsuno and Mendelzon 1991]. In these approaches, it is assumed that an agent has a belief set $K$, consisting of a set of formulas that he believes. $K * A$ represents the agent's belief set after observing $A$. We can think of $Bel^0(\mathcal{S})$ as characterizing the agent's initial belief set $K$. For each proposition $A$, we can identify $Bel^1(A)$ with $K * A$. In this framework, we can also do *iterated* change: $Bel^n(A_1^{(1)} \wedge \ldots \wedge A_n^{(n)})$ is the agent's belief state after observing $A_1$, then $A_2, \ldots$, and then $A_n$.

As we have shown in previous work [1995a, 1994b], conditioning captures the intuition of *minimal change* that underlies most approaches to belief change. In particular, both belief revision and belief update can be viewed as instances of conditioning on the appropriate prior. As expected, the prior plausibility measures that correspond to revision as-sign plausibility only to runs where the system does not change, but are fairly unstructured in other respects. On the other hand, the prior plausibility measures that correspond to update allow the system to change states, but put other constraints on how change can occur. Roughly speaking, they prefer runs where surprising events occur as late as possible.

## 4  MARKOVIAN BELIEF CHANGE

As we said in the introduction, we would like a notion of belief change that allows us both to revise our previous beliefs about the world and to allow for a change in the world occurring. Moreover, we need to address the question of representing the plausibility measure on runs. Can we get measures with reasonable belief-change properties that can be represented in a natural and compact manner? In the probabilistic framework, the Markov assumption provides a solution to both problems. As we now show, it is also useful in the plausibilistic setting.

The definition of Markovian probability measures generalizes immediately to plausibility measures. A conditional plausibility space $(W, \mathcal{F}, D, \text{Pl})$ is *Markovian* if it satisfies the same conditions as in Definition 2.1, with Pr replaced by Pl. Given a Markovian plausibility space $(W, \mathcal{F}, D, \text{Pl})$, we define the *transition plausibilities* analogously to the transition plausibilities: that is, $t_{s,s'} = \text{Pl}(S_{n+1} = s' \mid S_n = s)$.

In the probabilistic setting, the Markov assumption has many implications that can be exploited. In particular, we can easily show the existence and uniqueness of a Markovian prior with a given set of transition probabilities. Can we get a similar result for Markovian plausibility spaces? In general, the answer is no. To get this property, and the other desirable properties of Markovian plausibility spaces, we need to put more structure on plausibility measures.

In showing that there is a unique Markovian measure determined by the transition probabilities, we made use of two important properties of probability: The first is that $\Pr(A, B)$ is determined as $\Pr(B|A) \times \Pr(A)$. (We used this to calculate $\Pr([s_0, \ldots, s_n])$.) The second is that $\Pr(A \cup B) = \Pr(A) + \Pr(B)$, if $A$ and $B$ are disjoint. (This was used to get the probability of an arbitrary time-$n$ event from the probability of the time-$n$ prefixes.) We would like to have analogues of $+$ and $\times$ for plausibility.

To get an analogue of $+$, we need to assume that the plausibility of the union of two disjoint sets $A$ and $B$ is a function of the plausibility of $A$ and the plausibility of $B$. We call a plausibility measure Pl *decomposable* if there is a function $\oplus$ that is commutative, monotonic (so that if $d \leq d'$, then $d \oplus e \leq d' \oplus e$), and additive (so that $d \oplus \bot = d$ and $d \oplus \top = \top$) such that $\text{Pl}(A \cup B) = \text{Pl}(A) \oplus \text{Pl}(B)$ if $A$ and $B$ are disjoint. In [Friedman and Halpern 1995b], natural conditions are provided that guarantee that a plausibility measure is decomposable. Note that probability measures are decomposable, with $\oplus$ being $+$, possibility measures are decomposable with $\oplus$ being max, and $\kappa$-rankings are decomposable with $\oplus$ being min. By way of contrast, belief functions are not decomposable in general.

We next want to get an analogue of $\times$, to make sure that conditioning acts appropriately. More precisely, we want there to be a function $\otimes$ that is commutative, associative, strictly monotonic (so that if $d > d'$ and $e \neq \bot$, then $d \otimes e > d' \otimes e$), and bottom-preserving (so that $d \otimes \bot = \bot$)n such that $\text{Pl}(A|B, C) \otimes \text{Pl}(B|C) = \text{Pl}(A, B|C)$.

An *algebraic domain* is a tuple $\langle D, \oplus, \otimes \rangle$ where $D$ is a partially-ordered pointed domain, and $\oplus$ and $\otimes$ are binary operations on $D$ satisfying the requirements we described above, such that $\otimes$ distributes over $\oplus$. Algebraic domains are typically used in quantitative notions of uncertainty. For example, $\langle [0, 1], +, \times \rangle$ is used for probability and $\langle \mathbb{N}^*, \min, + \rangle$ is used for $\kappa$-rankings.[5] The standard examples of algebraic domains in the literature are totally ordered. However, it is not hard to construct partially-ordered algebraic domains. For example, consider $\langle [0, 1]^n, \oplus, \otimes \rangle$, where $\oplus$ and $\otimes$ are defined pointwise on sequences, and $\langle x_1, \ldots, x_n \rangle \preceq \langle y_1, \ldots, y_n \rangle$ if $x_i \leq y_i$ for $1 \leq i \leq n$. This is clearly a partially-ordered algebraic domain.

In this paper, we focus on plausibility spaces that are based on algebraic domains. A *structured* plausibility space $(W, \mathcal{F}, D, \text{Pl})$ is one for which there exist $\otimes$ and $\oplus$ such that $\langle D, \otimes, \oplus \rangle$, is an algebraic domain, and Pl is such that $\text{Pl}(A \cup B|E) = \text{Pl}(A|E) \oplus \text{Pl}(B|E)$ for disjoint $A$ and $B$, and $\text{Pl}(A|B, C) \otimes \text{Pl}(B|C) = \text{Pl}(A, B|C)$. (We remark that Darwiche [1992] and Weydert [1994] consider notions similar to structured plausibility spaces; we refer the reader to [Friedman and Halpern 1995b] for a more detailed discussion and comparison.)

From now on we assume that that Markovian plausibility spaces are structured. In the probabilistic setting, Markovian priors are useful in part because they can be described in a compact way. Similar arguments show that this is the case in the plausibilistic setting as well.

**Theorem 4.1:** *Let $\langle D, \oplus, \otimes \rangle$ be an algebraic domain, and let $\{t_{s,s'} : t_{s,s'} \in D\}$ be a set of transition plausibilities such that $\oplus_{s' \in S} t_{s,s'} = \top_D$ for all $s \in S$. Then there is a unique Markovian plausibility space $(W, \mathcal{F}, D, \text{Pl})$ such that $\text{Pl}(S_{n+1} = s' \mid S_n = s) = t_{s,s'}$ for all states $s$ and $s'$ and for all times $n$.*

**Proof:** (Sketch) Define Pl so that

$$\text{Pl}([s_0, \ldots, s_n]) = t_{s_0, s_1} \otimes \cdots \otimes t_{s_{n-1}, s_n}.$$

It is straightforward to show, using the properties of $\oplus$ and $\otimes$, that Pl is uniquely defined. ∎

Since we want to capture belief, we are particularly interested in qualitative plausibility spaces. Thus, it is of interest to identify when this construction results in a qualitative plausibility measure. It turns out that when the domain is totally ordered, we can ensure that the plausibility measure is qualitative by requiring $\oplus$ to be max.

---

[5]For possibility, $\langle \mathcal{R}, \max, \min \rangle$ is often used; this is not quite an algebraic domain, since min is not strictly monotonic. However, all our results but one—Theorem 5.3—holds as long as $\otimes$ is monotonic, even if it is not strictly monotonic.

**Proposition 4.2:** *Let $\langle D, \oplus, \otimes \rangle$ be an algebraic domain such that $\leq_D$ is totally ordered and $d \oplus d' = d$ whenever $d' \leq_D d$. Then the plausibility space of Theorem 4.1 is qualitative.*

It remains an open question to find natural sufficient conditions to guarantee that the plausibility space of Theorem 4.1 is qualitative when $D$ is only partially ordered.

We now have the tools to use the Markov assumption in belief change. To illustrate these notions, we examine how we would formalize Kautz's *stolen car* example [1986] and the variant discussed in the introduction.

**Example 4.3:** Recall that in the original story, the car is parked at time 0 and at time 3 the owner returns to find it gone. In the variant, the car is parked with a full fuel tank at time 0, at time 2 the owner returns to find it parked where he left it, but at time 3 he observes that the fuel tank is empty. To model these examples we assume there are three states: $s_{p\bar{e}}$, $s_{pe}$, and $s_{\overline{pe}}$. In $s_{p\bar{e}}$, the car is parked with a full tank; in $s_{pe}$, the car is parked with an empty tank; and in $s_{\overline{pe}}$ the car is not parked and the tank is full. We consider two propositions: $Parked = \{s_{p\bar{e}}, s_{pe}\}$ and $Full = \{s_{p\bar{e}}, s_{\overline{pe}}\}$.

In the original story, the evidence at time 3 is captured by $E_{\text{stolen}} = Parked^{(0)} \cap Full^{(0)} \cap \overline{Parked}^{(3)}$, since $Parked \cap Full$ is observed at time 0, and $\overline{Parked}$ is observed at time 3. Similarly, the evidence in the variant at time 2 is captured by $E^2_{\text{borrowed}} = Parked^{(0)} \cap Full^{(0)} \cap Parked^{(2)}$, and at time 3 by $E^3_{\text{borrowed}} = E^2_{\text{borrowed}} \cap Parked^{(3)} \cap \overline{Full}^{(3)}$.

We now examine one possible Markovian prior for this system. The story suggests that the most likely transitions are the ones where no change occurs. Suppose we attempt to capture this using $\kappa$-rankings. Recall that $\kappa$-rankings are based on the algebraic domain $\langle I\!N^*, \min, + \rangle$. We could, for example, take $t_{s_{p\bar{e}}, s_{p\bar{e}}} = t_{s_{pe}, s_{pe}} = t_{s_{\overline{pe}}, s_{\overline{pe}}} = 0$. If we believe that the transition from $s_{p\bar{e}}$ to $s_{pe}$ is less likely than the transition from $s_{p\bar{e}}$ to $s_{\overline{pe}}$ and from $s_{\overline{pe}}$ to $s_{pe}$, we can take $t_{s_{p\bar{e}}, s_{pe}} = 3, t_{s_{p\bar{e}}, s_{\overline{pe}}} = 1$, and $t_{s_{\overline{pe}}, s_{pe}} = 1$.

Suppose we get the evidence $E_{\text{stolen}}$, that the car is parked at time 0 but gone at time 3. It is easy to verify that there are exactly three 3-prefixes with plausibility 0 after we condition on $E_{\text{stolen}}$: $[s_{p\bar{e}}, s_{\overline{pe}}, s_{\overline{pe}}, s_{\overline{pe}}]$, $[s_{p\bar{e}}, s_{p\bar{e}}, s_{\overline{pe}}, s_{\overline{pe}}]$ and $[s_{p\bar{e}}, s_{p\bar{e}}, s_{p\bar{e}}, s_{\overline{pe}}]$. Thus, the agent believes that the car was stolen before time 3, but has no more specific beliefs as to when.

Suppose we instead get the evidence $E^2_{\text{borrowed}}$, i.e., that the car is parked and has a full tank at time 0, and is still parked at time 2. In this case, the most plausible time-2 prefix is $[s_{p\bar{e}}, s_{p\bar{e}}, s_{p\bar{e}}]$; the expected observation that the car is parked does not cause the agent to believe that any change occurred. What happens when he observes that the tank is empty, i.e., $E^3_{\text{borrowed}}$? There are two possible explanations: Either the gas leaked at some stage (so that there was a transition from $s_{p\bar{e}}$ to $s_{pe}$ before time 3) or the car was "borrowed" without the agent's knowledge (so that there was a transition from $s_{p\bar{e}}$ to $s_{\overline{pe}}$ and then from $s_{\overline{pe}}$ to $s_{pe}$). Applying the definition of $\otimes$ we conclude that $\text{Pl}([s_{p\bar{e}}, s_{\overline{pe}}, s_{pe}]) = 2$ and $\text{Pl}([s_{p\bar{e}}, s_{p\bar{e}}, s_{pe}]) = 3$. Thus, the agent considers the most likely explanation to be that the car was borrowed.

It is worth noticing how the agent's beliefs after observing $E^3_{\text{borrowed}}$ depend on details of the transition plausibilities and the $\otimes$ function. For the $\kappa$-ranking above, we have $t_{s_{p\bar{e}}, s_{pe}} > t_{s_{p\bar{e}}, s_{\overline{pe}}} \otimes t_{s_{\overline{pe}}, s_{pe}}$; with this choice, the most likely explanation is $[s_{p\bar{e}}, s_{\overline{pe}}, s_{pe}]$. If, instead, we had used a $\kappa$-ranking such that $t_{s_{p\bar{e}}, s_{pe}} < t_{s_{p\bar{e}}, s_{\overline{pe}}} \otimes t_{s_{\overline{pe}}, s_{pe}}$ (for example, by taking $t_{s_{p\bar{e}}, s_{pe}} = t_{s_{p\bar{e}}, s_{\overline{pe}}} = t_{s_{\overline{pe}}, s_{pe}} = 1$), then the most likely explanation would have been $\text{Pl}([s_{p\bar{e}}, s_{p\bar{e}}, s_{pe}])$. Finally, if we had used a $\kappa$-ranking such that $t_{s_{p\bar{e}}, s_{pe}} = t_{s_{p\bar{e}}, s_{\overline{pe}}} \otimes t_{s_{\overline{pe}}, s_{pe}}$ (for example, by taking $t_{s_{p\bar{e}}, s_{pe}} = 2$ and $t_{s_{p\bar{e}}, s_{\overline{pe}}} = t_{s_{\overline{pe}}, s_{pe}} = 1$), then the agent would have considered both explanations possible. ∎

As this example shows, using qualitative Markovian plausibilities, the agent can revise earlier beliefs as revision does (for example, the agent may revise his beliefs regarding whether the car was parked at time 1 once he sees that the fuel tank is no longer full), or he may think that a change occurred in the world that explains his current beliefs (the gas tank leaked). Of course, the agent might also consider both explanations to be likely.

It is interesting to compare the behavior of Markovian plausibility measures in this example to that of Katsuno and Mendelzon's update [1991]. As Kautz [1986] noted, given $E_{\text{stolen}}$, we should believe that the car was stolen during the period we were gone, but should not have more specific beliefs. Markovian measures give us precisely this conclusion. Update, on the other hand, leads us to believe that the car was stolen just before we notice that it is gone [Friedman and Halpern 1994b]. To see this, note that any observation that is implied by the agent's current beliefs does not change those beliefs (this is one of Katsuno and Mendelzon's postulates). Combined with the fact that update never revises beliefs about the past, we must conclude that the agent believes that the car was not stolen at time 1 or 2. In the second variant, the differences are even more significant. Using update, we conclude that there was a gas leak. Update cannot reach the conclusion that the car was borrowed, since that involves changing beliefs about the past: For the agent to consider it possible at time 3 that the car was borrowed, he has to change his belief that the car was parked at time 2. Moreover, update does not allow us to compare the relative plausibility of a gas leak to that of the car being borrowed. (See [Friedman and Halpern 1994b] for further discussion of update's behavior in this example.)

This discussion shows that Markovian priors are useful for representing plausibility measures compactly. Moreover, they give the user the right level of control: by setting the transition plausibilities, the user can encode his preference regarding various explanations. Markovian priors have computational advantages as well, as we now show.

Given $\otimes$ and the transition plausibilities, the Markov assumption allows us to compute the plausibilities of every $n$-prefix; then using $\oplus$, we can compute whether an event $A$ is in $Bel^n_{Pl}(E)$. When $n$ is large, this procedure is unreasonable. We do not want to examine all $n$-prefixes in order to evaluate our beliefs about time $n$. Fortunately, the Markov assumption allows us to maintain our beliefs about the current state of the system in an efficient way.

Suppose that the agent makes a sequence of observations $O_1, O_2, \ldots$. Each observation is a proposition (i.e., a set of states). The evidence at time $n$ is simply the accumulated evidence: $E_n = O_1^{(1)} \cap \ldots \cap O_n^{(n)}$. We are interested in testing whether $A^{(n)} \in Bel^n_{Pl}(E_n)$. According to C1, this is equivalent to testing whether $Pl(A^{(n)}, E_n) > Pl(\overline{A^{(n)}}, E_n)$.

It is easy to see that the plausibilities $Pl(S_n = s, E_n)$, $s \in \mathcal{S}$, suffice for determining whether the agent believes $A$. This follows from the observation that

$$Pl(A^{(n)}, E_n) = \oplus_{s \in A} Pl(S_n = s, E_n),$$

and similarly for $Pl(\overline{A}^{(n)}, E_n)$.

In addition, it is straightforward for an agent who has the plausibilities $Pl(S_n = s, E_n)$, $s \in \mathcal{S}$, to update them when he makes a new observation. To see this, observe that

$$Pl(S_{n+1} = s, E_{n+1}) = Pl(S_{n+1} = s, E_n, O_{n+1}^{(n+1)}).$$

Thus, $Pl(S_{n+1} = s, E_{n+1}) = \bot$ if $s \notin O_{n+1}$; otherwise, it is $Pl(S_{n+1} = s, E_n)$. In algebraic domains, we can compute the latter plausibility using much the same techniques as in probabilistic domains:

$$\begin{aligned} & Pl(S_{n+1} = s \cap E_n) \\ = \ & \oplus_{s' \in S} Pl(S_{n+1} = s | S_n = s', E_n) \otimes Pl(S_n = s', E_n) \end{aligned}$$

Using the Markov assumption, $Pl(S_{n+1} = s | S_n = s', E_n) = Pl(S_{n+1} = s | S_n = s') = t_{s',s}$. Thus, we can compute $Pl(S_{n+1} = s, E_{n+1})$ using $Pl(S_n = s', E_n)$ and the transition plausibilities $t_{s',s}$.

To summarize, there is a simple two-stage procedure for updating beliefs given new observations. In the first stage, we compute the plausibility of the current states using our beliefs about the previous state. In the second stage, we prune states that are inconsistent with the current observation.

Again, it is interesting to compare this approach to Katsuno and Mendelzon's belief update. One of the assumptions made by update is that the agent maintains only his beliefs about the current state of the system. Roughly speaking, this amounts to tracking only the states of the system that have maximal plausibility given past observations. Thus, update can require less information to update beliefs. This, however, comes at the price of assuming that abnormalities are delayed as much as possible. This, as we saw in the case of the borrowed car example, may lead to unintuitive conclusions. We conjecture that to avoid such conclusions, agent must keep track of either information about past events or degrees of plausibilities of all possible states of the system at the current time.

### 4.1 Characterizing Markovian Belief Change

Our formalization of belief change is quite different from most in the literature. Most approaches to belief change start with a collection of postulates, argue that they are reasonable, and prove some consequences of these postulates. Occasionally, a semantic model for the postulates is provided and a representation theorem is proved (of the form that every semantic model corresponds to some belief revision process, and that every belief revision process can be captured by some semantic model).

We have not discussed postulates at all here. There is a good reason for this: Markovian belief change does not satisfy any postulates of the standard sort beyond those of any other notion of belief change. To make this precise, assume that we have some language $\mathcal{L}$ that includes the usual propositional constructs (i.e., $\vee, \wedge, \neg$ and $\Rightarrow$). A *plausibility structure* (over $\mathcal{L}$) is a tuple $PL = (W, \mathcal{F}, D, Pl, \pi)$, where $(W, \mathcal{F}, D, Pl)$ is a conditional plausibility space, and $\pi$ is a truth assignment that associates with each world in $W$ a complete consistent set of formulas in $\mathcal{L}$. Given a plausibility structure, we can associate with each formula $\varphi$ in $\mathcal{L}$ the set of states where $\varphi$ is true: $[\![\varphi]\!]_{PL} = \{s \in \mathcal{S} : \pi(s)(\varphi) = \mathbf{true}\}$. Observing the sequence $\varphi_1, \ldots, \varphi_n$ at times $1, \ldots, n$ amounts to conditioning on the event $E_{PL, \varphi_1, \ldots, \varphi_n} = [\![\varphi_1]\!]_{PL}^{(1)} \cap \ldots \cap [\![\varphi_n]\!]_{PL}^{(n)}$. Similarly, $\varphi$ is believed at time $n$, given $E$, if $[\![\varphi]\!]_{PL} \in Bel^n_{Pl}(E)$.

We now define two classes of structures. Let $\mathcal{P}^{QPL}$ consist of all qualitative plausibility structures, and let $\mathcal{P}^{QPL,M}$ consist of all qualitative plausibility structures with a Markovian plausibility measure. Since $\mathcal{P}^{QPL,M} \subset \mathcal{P}^{QPL}$, any formula valid in $\mathcal{P}^{QPL}$ (that is, true in every plausibility structure in $\mathcal{P}^{QPL}$) must also be valid in $\mathcal{P}^{QPL,M}$. As we now show, the converse holds as well.

**Theorem 4.4:** *Let $PL = (W, \mathcal{F}, D, Pl, \pi)$ be a plausibility structure in $\mathcal{P}^{QPL}$ and let $n > 0$. Then there is a plausibility structure $PL' = (W', \mathcal{F}', D', Pl', \pi')$ in $\mathcal{P}^{QPL,M}$ such that for all sequences of formulas $\varphi_1, \ldots, \varphi_m$, $m \leq n$ and all formulas $\psi$, $[\![\psi]\!]_{PL} \in Bel^m_{Pl}(E_{PL, \varphi_1, \ldots, \varphi_m})$ if and only if $[\![\psi]\!]_{PL'} \in Bel^m_{Pl'}(E_{PL', \varphi_1, \ldots, \varphi_m})$.*

**Proof:** (Sketch) Define $\mathcal{S}' = \{\langle s_1, \ldots, s_m \rangle : s_1, \ldots, s_m \in \mathcal{S}, m \leq n\}$ and $\pi'(\langle s_1, \ldots, s_m \rangle) = \pi(s_m)$. We can then construct a Markovian plausibility measure over $W'$ that simulates Pl up to time $n$, in that $Pl'([\langle s_0 \rangle, \langle s_0, s_1 \rangle, \ldots, \langle s_0, s_1, \ldots, s_m \rangle]) = Pl([s_0, s_1, \ldots, s_m])$ for all $m \leq n$. ∎

It follows from Theorem 4.4 that for any Markovian plausibility structure $PL_1$, there is a non-Markovian plausibility structure $PL_2$ such that, for every sequence $\varphi_1, \ldots, \varphi_n$ of observations in $\mathcal{L}$, the agent has the same beliefs (in the language $\mathcal{L}$) after observing $\varphi_1, \ldots, \varphi_n$, no matter whether his plausibility is characterized by $PL_1$ or $PL_2$. Thus, there are no special postulates for Markovian belief change over and above the postulates that hold for any approach to belief change.

## 5  PARTIALLY SPECIFIED TRANSITIONS

Up to now, we have implicitly assumed that when the user specifies transition plausibilities, he has some underlying algebraic domain in mind. This assumption is quite strong, since it assumes that the user knows the exact plausibility value of the transitions, and the functions $\otimes$ and $\oplus$ that relate them.

In this section, we focus on situations where the user only specifies a partial order on transitions. This is a natural form of knowledge that can be specified/assessed in a relatively straightforward manner. Given a set $S$ of states, consider *transition variables* of the form $x_{s,s'}$, for $s, s' \in S$. We think of $x_{s,s'}$ as a variable representing the plausibility of the transition from $s$ to $s'$. Assume that we are given constraints on these transition variables, specified by a partial order $\leq_t$ on the variables, together with constraints of the form $x_{s,s'} \leq_t \bot$.[6] These can be thought of as constraints on the relative plausibility of transitions, together with constraints saying that some transitions are impossible. $x_{s,s'} = \bot$. We are interested in plausibility measures that are consistent with sets of such constraints.

We say that a set $C$ of constraints is *safe* if there is no variable $y$ and state $s$ such that the $x_{s,s'} <_t y$ is in $C$ for all $s' \in S$. There is no qualitative plausibility measure that satisfies an unsafe set of constraints. To see this, note that, for a given $s$, $t_{s,s'}$ represent the plausibilities of disjoint events for different choices of $s'$. If we are dealing with qualitative plausibility, if $t_{s,s'} < t$ for all $s'$, then $\bigoplus_{s' \in S} t_{s,s'} < t$. However, we also have $\bigoplus_{s' \in S} t_{s,s'} = \top$, and we cannot have $t > \top$. Thus, no qualitative plausibility measure can satisfy an unsafe set of constraints. However, any safe set set of constraints is satisfiable.

**Theorem 5.1:** *Given a safe set $C$ of constraints on the transitions over $S$ as above, there is a qualitative Markovian plausibility space $(W, \mathcal{F}, D, \text{Pl})$ consistent with $C$.*

Note that this theorem guarantees the existence of a *qualitative* prior. Unlike the situation described in the previous section, where transition plausibilities were fully specified, there is not in general a unique (qualitative) plausibility measure Pl consistent with $\leq_t$. This is due to the fact that we can choose various plausibility values as well as various operators $\otimes$ and $\oplus$ in the algebraic domain, while remaining consistent with $\leq_t$. To see this, consider the following variant of Example 4.3.

**Example 5.2:** Using the notation of Example 4.3, consider the constraints $x_{s_{p\bar{e}},s_{pe}} <_t x_{s_{p\bar{e}},s_{\overline{pe}}} =_t x_{s_{\overline{p}e},s_{pe}} <_t x_{s_{p\bar{e}},s_{p\bar{e}}} =_t x_{s_{\overline{pe}},s_{\overline{pe}}} =_t x_{s_{pe},s_{pe}}$. One way of satisfying these constraints is by using the standard $\kappa$-ranking described in Example 4.3, for which we have $t_{s_{p\bar{e}},s_{pe}} = 3$, $t_{s_{p\bar{e}},s_{\overline{pe}}} = t_{s_{\overline{pe}},s_{pe}} = 1$, and $t_{s,s} = 0$ for all states $s$. If Pl is the plausibility function generated by this $\kappa$-ranking, we have $\text{Pl}([s_{p\bar{e}}, s_{p\bar{e}}, s_{pe}]) = 3$ and $\text{Pl}([s_{p\bar{e}}, s_{\overline{pe}}, s_{pe}]) = 2$, since $\otimes$ is $+$ for $\kappa$-rankings.

However, now consider another way of satisfying the same constraints, again using $\kappa$-rankings. Let $\text{Pl}'$ be a Markovian plausibility measure with the following transition plausibilities: $t'_{s_{p\bar{e}},s_{pe}} = 3$, $t'_{s_{p\bar{e}},s_{\overline{pe}}} = t'_{s_{\overline{pe}},s_{pe}} = 2$, and $t'_{s,s} = 0$ for all states $s$. It is easy to verify that $\text{Pl}'$ satisfies the constraints we described above. However, it is easy to check that we have $\text{Pl}'([s_{p\bar{e}}, s_{\overline{pe}}, s_{pe}]) = 4$ and $\text{Pl}'([s_{p\bar{e}}, s_{p\bar{e}}, s_{pe}]) = 3$. This means that $\text{Pl}([s_{p\bar{e}}, s_{p\bar{e}}, s_{pe}]) > \text{Pl}'([s_{p\bar{e}}, s_{\overline{pe}}, s_{pe}])$, while $\text{Pl}'([s_{p\bar{e}}, s_{p\bar{e}}, s_{pe}]) < \text{Pl}([s_{p\bar{e}}, s_{\overline{pe}}, s_{pe}])$. Thus, although Pl and $\text{Pl}'$ satisfy the same constraints on transition plausibilities, they lead to different orderings of the 3-prefixes, and thus to different notions of belief change. ∎

Constraints on the relative order of transition plausibilities do give rise to some constraints on the relative plausibility of $n$-prefixes. It is these constraints that we now study. We start with some notation. Let $C$ be a safe set of constraints on transition variables, and let Pl be a Markovian measure consistent with these constraints. As we observed, $\text{Pl}([s_0, \ldots, s_n]) = t_{s_0,s_1} \otimes \cdots \otimes t_{s_{n-1},s_n}$ once we fix the transition plausibilities and $\otimes$. Thus, the plausibility of a sequence is determined by the transitions involved. The constraints in $C$ determine an ordering $\preceq$ on $n$-prefixes as follows:

$[s_0, s_1, \ldots, s_n] \preceq [s_0, s'_1, \ldots, s'_n]$ if there is some constraint of the form $x_{s'_i, s'_{i+1}} = \bot$ in $C$ (where we take $s'_0 = s_0$), or if there is some permutation $\sigma$ over $\{0, 1, \ldots, n\}$ such that $\sigma(0) = 0$ and $x_{s_i,s_{i+1}} \leq_t x_{s_{\sigma(i)},\sigma(i+1)}$ is in $C$, for $i = 0, \ldots, n-1$.[7]

We define $\prec$ and $\approx$ using $\preceq$ in the standard manner. If $[s_0, s_1, \ldots, s_n] \approx [s_0, s'_1, \ldots, s'_n]$ we say these two $n$-prefixes are *equivalent*. Intuitively, $\preceq$ captures what is forced by all Markov plausibility measures consistent with $\leq_t$, since it captures what is true for all choices of $\otimes$ in normal conditional plausibility spaces.

**Theorem 5.3:** *Let Pl be a Markovian plausibility measure consistent with some constraints $C$, and let $\preceq$ be defined as above in terms of $C$. If $[s_0, s_1, \ldots, s_n] \preceq [s_0, s'_1, \ldots, s'_n]$, then $\text{Pl}([s_0, s_1, \ldots, s_n]) \leq \text{Pl}([s_0, s'_1, \ldots, s'_n])$. Moreover, if $[s_0, s_1, \ldots, s_n] \prec [s_0, s'_1, \ldots, s'_n]$, then $\text{Pl}([s_0, s_1, \ldots, s_n]) < \text{Pl}([s_0, s'_1, \ldots, s'_n])$.*

The converse to Theorem 5.3 does not hold in general. A Markovian plausibility measure consistent with $C$ may introduce more comparisons between $n$-prefixes than those determined by the $\preceq$ ordering.

**Example 5.4:** Consider again the setup of Example 4.3, using the constraints on transitions considered in Example 5.2. Applying the definition of $\preceq$, we find that $[s_{p\bar{e}}, s_{p\bar{e}}, s_{pe}] \prec$ $[s_{p\bar{e}}, s_{p\bar{e}}, s_{p\bar{e}}]$ and similarly $[s_{p\bar{e}}, s_{\overline{pe}}, s_{pe}] \prec [s_{p\bar{e}}, s_{p\bar{e}}, s_{p\bar{e}}]$. It is easy to verify that the plausibility measure Pl described in Example 5.2 satisfies these constraints. However, $[s_{p\bar{e}}, s_{p\bar{e}}, s_{pe}]$ is incomparable to $[s_{p\bar{e}}, s_{\overline{pe}}, s_{pe}]$ according to $\preceq$, but, as we calculated in Example 5.2, we have $\text{Pl}([s_{p\bar{e}}, s_{p\bar{e}}, s_{pe}]) < \text{Pl}([s_{p\bar{e}}, s_{\overline{pe}}, s_{pe}])$. ∎

---

[6] Of course, we take $x =_t y$ to be an abbreviation for $x \leq_t y$ and $y \leq_t x$, and $x <_t y$ to be an abbreviation for $x \leq_{ty}$ and not $(y \leq_t x)$.

[7] Note that we take $s_0 = s'_0$ and $\sigma(0) = 0$ because we are only interested in $n$-prefixes with initial state $s_0$.

Although, in general, a Markovian plausibility measure will place more constraints than those implied by $\preceq$, we can use the construction of Theorem 5.1 to show that there is in fact a plausibility measure that precisely captures $\preceq$.

**Theorem 5.5:** *Given a safe set of constraints $C$, there is a qualitative Markovian plausibility measure* $\text{Pl}_*$ *consistent with $C$ such that $[s_0, s_1, \ldots, s_n] \preceq [s_0, s'_1, \ldots, s'_n]$ if and only if* $\text{Pl}_*([s_0, s_1, \ldots, s_n]) \leq \text{Pl}_*([s_0, s'_1, \ldots, s'_n])$.

These results show that examining the relative plausibility of transitions allows to deduce the relative plausibilities of some $n$-prefixes. We can use this knowledge to conclude what beliefs must hold in any Markovian measure with these transition plausibilities. We need an additional definition:

$MAX^n(E) =_{\text{def}} \bigcup \{[s_0, s_1, \ldots, s_n] \subseteq E \mid$
$\forall [s_0, s'_1, \ldots, s'_n] \subseteq E, [s_0, s_1, \ldots, s_n] \not\prec [s_0, s'_1, \ldots, s'_n]\}$.

$MAX^n(E)$ is the event defined as the union of $n$-prefixes in $E$ that are maximal according to $\prec$. It easily follows from axiom A2 that in any qualitative Markovian measure consistent with $\leq_t$, the plausibility of $MAX^n(E)$ given $E$ is greater than the plausibility of $\overline{MAX^n(E)}$ given $E$. As a consequence, we have the following result.

**Theorem 5.6:** *Suppose Pl is a qualitative Markovian measure consistent with some set $C$ of constraints, $E$ is a time-$n$ event, and $A \subseteq S$. If $MAX^n(E) \subseteq A^{(n)}$, then $A \in Bel^n(E)$.*

Thus, by examining the most plausible (according to $\preceq$) $n$-prefixes, we get a sufficient condition for a set $A$ to be believed. The converse to Theorem 5.6 does not hold: the agent might believe $A$ even if some of the $n$-prefixes are not in $A^{(n)}$. However, the $n$-prefixes in $MAX^n(E)$ are equally plausible, then the converse does hold.

**Proposition 5.7:** *Suppose Pl is a qualitative Markovian measure consistent with some set $C$ of constraints, $E$ is a time-$n$ event, and $A \subseteq S$. If all the $n$-prefixes in (i.e., the ones that are subsets of) $MAX^n(E)$ are equivalent, then $MAX^n(E) \subseteq A^{(n)}$ if and only if $A \in Bel^n(E)$.*

**Example 5.8:** We now examine the setup of Example 4.3 using partially specified transition plausibilities. To capture our intuition that changes are unlikely, we require that $x_{s,s} \leq_t x_{s',s''}$ for all $s, s', s'' \in \{s_{p\bar{e}}, s_{pe}, s_{\overline{pe}}\}$. We also assume that $x_{s_{p\bar{e}}, s_{pe}}$, $x_{s_{p\bar{e}}, s_{\overline{pe}}}$, and $x_{s_{\overline{pe}}, s_{pe}}$ are all strictly less likely than $x_{s_{p\bar{e}}, s_{p\bar{e}}}$, but are not comparable to each other. Finally, we assume all other transitions are impossible.

Suppose we get the evidence $E_{\text{stolen}}$, that the car is parked at time 0 but gone at time 3. It is easy to verify that $MAX^3(E_{\text{stolen}})$ consists of the three 3-prefixes $[s_{p\bar{e}}, s_{\overline{pe}}, s_{\overline{pe}}, s_{\overline{pe}}]$, $[s_{p\bar{e}}, s_{p\bar{e}}, s_{\overline{pe}}, s_{\overline{pe}}]$, and $[s_{p\bar{e}}, s_{p\bar{e}}, s_{p\bar{e}}, s_{\overline{pe}}]$. Moreover, it easy to check that these three 3-prefixes are equivalent. From Proposition 5.7, it now follows that if Pl is a qualitative Markovian measure, then $MAX^3(E_{\text{stolen}})$ characterizes the agent's beliefs. The agent believes that the car was stolen before time 3, but has no more specific beliefs as to when. This proves that all Markovian priors that are consistent with $\leq_t$ generate the same beliefs as the $\kappa$-ranking described in Example 5.2, given this observation.

Suppose we instead get the evidence $E^2_{\text{borrowed}}$, that the car is parked and has a full tank at time 0, and is still parked at time 2. In this case, we have $MAX^2(E^2_{\text{borrowed}}) = [s_{p\bar{e}}, s_{p\bar{e}}, s_{p\bar{e}}]$—the expected observation that the car is parked does not cause the agent to believe that any change occurred. What happens when he observes that the tank is empty, i.e., $E^3_{\text{borrowed}}$? As noted in Example 4.3, there are two possible explanations: Either the gas leaked or the car was "borrowed". Without providing more information, we should not expect the agent to consider one of these cases to be more plausible than the others. Indeed, it is easy to verify that $MAX^3(E^3_{\text{borrowed}})$ contains all the 3-prefixes that correspond to these two explanations. Unlike the previous case, some of these 3-prefixes are not equivalent. Thus, different qualitative Markovian plausibility measures may lead to different beliefs at time 3, even if they are consistent with the specifications.

If we add the further constraint $x_{s_{p\bar{e}}, s_{\overline{pe}}} <_t x_{s_{p\bar{e}}, s_{pe}}$ (so that a gas leak is more plausible than a theft), then the most likely 3-prefixes are the ones where there is a gas leak. However, if we specify that $x_{s_{p\bar{e}}, s_{pe}}$ is less likely than both $x_{s_{p\bar{e}}, s_{\overline{pe}}}$ and $x_{s_{\overline{pe}}, s_{pe}}$ (so that a gas leak is less plausible than the car being taken or returned), then $MAX^3(E^3_{\text{borrowed}})$ will still contain both explanations. Even with this additional information, $\preceq$ cannot compare $[s_{p\bar{e}}, s_{\overline{pe}}, s_{pe}, s_{pe}]$ to $[s_{p\bar{e}}, s_{pe}, s_{pe}, s_{pe}]$, because although $t_{s_{p\bar{e}}, s_{pe}} < t_{s_{p\bar{e}}, s_{\overline{pe}}}$ and $t_{s_{p\bar{e}}, s_{pe}} < t_{s_{\overline{pe}}, s_{pe}}$, it might be that $t_{s_{p\bar{e}}, s_{pe}} \otimes t_{s_{pe}, s_{p\bar{e}}} \not< t_{s_{p\bar{e}}, s_{\overline{pe}}} \otimes t_{s_{\overline{pe}}, s_{pe}}$. Our specification, in general, does not guarantee that one explanation is preferred to the other. ∎

We note that we can use the procedure described above to maintain an estimate of the agents beliefs at each time point. This involves using the Markovian plausibility space $\text{Pl}_*$ of Theorem 5.5. We discuss the details in the full paper.

# 6 RELATED WORK

We now briefly compare our approach to others in the literature.

Markovian belief change provides an approach for dealing with sequences of observations. *Iterated* belief revision, which also deals with sequences of observations, has been the focus of much recent attention (see [Lehmann 1995] and the references therein). Conditioning a prior plausibility measure provides a general approach to dealing with iterated belief revision. By using conditioning, we are implicitly assuming that the observations made provide correct information about the world. We cannot condition on an inconsistent sequence of observations. This assumption allows us to avoid some of the most problematic aspects of belief revision, and focus our attention instead on putting additional structure into the prior, so as to be able to express in a straightforward way natural notions of belief change.

One of the goals of Markovian belief change is to be able to combine aspects of revision and update. Recent work of Boutilier [1995] is similarly motivated. Essentially, Boutilier proposes conditioning on sequences of length two, using $\kappa$-rankings. While he does not pursue the probabilistic analogy, his discussion describes the belief change operation as a combination of beliefs about the state of the system, i.e., $\text{Pl}(S_1 = s)$ and beliefs about the likelihood of transitions, i.e., $\text{Pl}(S_2 = s'|S_1 = s)$. Boutilier proposes a two-stage procedure for updating belief which is similar to the one we outlined in Section 4. It is important to note that we derived this procedure using standard probabilistic arguments, something that Boutilier was not able to do in his framework. Our work, which was done independently of Boutilier's, can be viewed as extending his framework. We have arbitrary sequences of states, not just one-step transitions. In addition, because our approach is based on plausibility, not $\kappa$-rankings, we can deal with partially-ordered plausibilities.

Finally, we note that the Markov assumption has been used extensively in the literature on probabilistic reasoning about action. Papers on this topic typically represent situations using *dynamic belief networks* [Dean and Kanazawa 1989]. Dynamic belief networks are essentially Markov chains with additional structure on states: A state is assumed to be characterized by the values of a number of variables. The probability of a transition is described by a belief network. Belief networks allow us to express more independence assumptions than just those characterizing Markovian probabilities. For example, using a belief network, we can state that the value of variable $X_2$ at time $n + 1$ is independent of the value of $X_1$ at time $n$ given the value of $X_2$ at time $n$. Darwiche [1992] showed how qualitative Bayesian networks could be captured in his framework; it should be straightforward to add such structure to the plausibility framework as well, once we restrict to structured plausibility spaces.

## 7 DISCUSSION

This paper makes two important contributions. First, we demonstrate how the Markov assumption can be implemented in a qualitative setting that allows for a natural notion of belief. While similar intuitions regarding independence may have guided previous work on belief change (e.g., [Katsuno and Mendelzon 1991]), we are the first to make an explicit Markovian assumption. We believe that this approach provides a natural means of constructing a plausibility assessment over sequences given an assessment of the plausibility of transitions. Moreover, as we have seen, this assumption also leads to computational benefits similar to these found in the probabilistic setting.

Second, we examined what conclusions can be drawn given only an ordering on the transition plausibilities. Starting with a specification of constraints on the relative plausibility of transitions, we describe properties of the belief change operation. Since all we consider are comparisons between plausibilities of transitions, our conclusions are not always that strong. Of course, it is reasonable to consider richer forms of constraints, that might also constrain the behavior of $\otimes$ and $\oplus$. With more constraints, we can typically prove more about how the agent's beliefs will change. In future work, we plan to examine the consequences of using richer constraints.

One of the standard themes of the belief change literature is the effort to characterize a class of belief-change operators by postulates, such as the AGM postulates for belief revision [Alchourrón, Gärdenfors, and Makinson 1985] and the Katsuno-Mendelzon postulates for belief update [Katsuno and Mendelzon 1991]. As the discussion in Section 4.1 shows, characterizing Markovian belief change with such postulates does not provide additional insight. The trouble is that such postulates do *not* let us reason about relative likelihoods of transitions and independence of transitions. But this is precisely the kind of reasoning that motivates the use of Markovian plausibility (and probability) measures. Such reasoning clearly plays a key role in modeling the stolen car problem and its variants. If the language is too weak to allow us to express such reasoning (as is the case for the language used to express the AGM postulates), then we cannot distinguish the class of Markovian plausibility measures from the class of all plausibility measures.

This suggests that the right language to reason about belief change should allow us to talk about transitions and their relative plausibility. As our examples show, applications are often best described in these terms. This observation has implications for reasoning about action as well as belief change. Typical action-representation methodologies, such as the situation calculus [McCarthy and Hayes 1969], describe the changes each action brings about when applied in various states. Such a description talks about the *most likely* transition, given that the action occurred. Most approaches, however, do not explicitly deal with the less likely effects of actions. In a certain class of problems (called *predictive* problems in [Lin and Shoham 1991]), it suffices to specify only the most likely transitions. Roughly speaking, in these problems we are given information about the initial state, and we are interested in the beliefs about the state of the world after some sequence of actions. In such theories, we *never* get a surprising observation. Thus, only the most likely transitions play a role in the reasoning process. Problems appear when surprising observations are allowed. Kautz's stolen car problem is a canonical example of a situation with a surprising observation. As our example illustrates, in order to get reasonable conclusions in such theories, we need to provide information about the relative likelihood of all transitions, instead of identifying just the most likely transitions. We believe that a language for reasoning about actions and/or beliefs that change over time should allow the user to compare the plausibility of transitions (and sequences of transitions), giving as much or as little detail as desired. We believe that our approach provides the right tools to deal with these issues. We hope to return to this issue in future work.

### Acknowledgements

The authors are grateful to Craig Boutilier, Ronen Brafman, Moises Goldszmidt, and Daphne Koller, for comments on drafts of this paper and useful discussions relating to this work. This work was supported in part by NSF Grant IRI-95-03109. The first author was also supported in part by IBM Graduate Fellowship.

## References

Alchourrón, C. E., P. Gärdenfors, and D. Makinson (1985). On the logic of theory change: partial meet functions for contraction and revision. *Journal of Symbolic Logic 50*, 510–530.

Boutilier, C. (1995). Generalized update: belief change in dynamic settings. In *IJCAI '95*, pp. 1550–1556.

Darwiche, A. (1992). *A Symbolic Generalization of Probability Theory*. Ph. D. thesis, Stanford University.

Dean, T. and K. Kanazawa (1989). A model for reasoning about persistence and causation. *Computational Intelligence 5*, 142–150.

Dubois, D. and H. Prade (1990). An introduction to possibilistic and fuzzy logics. In *Readings in Uncertain Reasoning*, pp. 742–761. Morgan Kaufmann.

Dubois, D. and H. Prade (1995). Numerical representations of acceptance. In *UAI '95*, pp. 149–156.

Friedman, N. and J. Y. Halpern (1994b). A knowledge-based framework for belief change. Part II: revision and update. In *KR '94*, pp. 190–201.

Friedman, N. and J. Y. Halpern (1995a). Modeling belief in dynamic systems. part I: foundations. Technical Report RJ 9965, IBM. Submitted for publication. A preliminary version appears in R. Fagin editor. *Theoretical Aspects of Reasoning about Knowledge: Proc. Fifth Conference*, 1994, pp. 44–64, under the title "A knowledge-based framework for belief change. Part I: foundations".

Friedman, N. and J. Y. Halpern (1995b). Plausibility measures: a user's manual. In *UAI '95*, pp. 175–184.

Friedman, N. and J. Y. Halpern (1996). Plausibility measures and default reasoning. In *AAAI '96*. Extended version appears as IBM Technical Report RJ 9959, 1995.

Gärdenfors, P. (1988). *Knowledge in Flux*. MIT Press.

Goldszmidt, M. and J. Pearl (1992). Rank-based systems: A simple approach to belief revision, belief update and reasoning about evidence and actions. In *KR '92*, pp. 661–672.

Howard, R. A. (1971). *Dynamic Probabilistic Systems. Volume I: Markov Models*. Wiley.

Katsuno, H. and A. Mendelzon (1991). On the difference between updating a knowledge base and revising it. In *KR '91*, pp. 387–394.

Kautz, H. A. (1986). Logic of persistence. In *AAAI '86*, pp. 401–405.

Kemeny, J. G. and J. L. Snell (1960). *Finite Markov Chains*. Van Nostrand.

Kraus, S., D. Lehmann, and M. Magidor (1990). Nonmonotonic reasoning, preferential models and cumulative logics. *Artificial Intelligence 44*, 167–207.

Lehmann, D. (1995). Belief revision, revised. In *IJCAI '95*, pp. 1534–1540.

Lin, F. and Y. Shoham (1991). Provably correct theories of action (preliminary report). In *AAAI '91*, pp. 349–354.

McCarthy, J. and P. J. Hayes (1969). Some philosophical problems from the standpoint of artificial intelligence. In D. Michie (Ed.), *Machine Intelligence 4*, pp. 463–502.

Pearl, J. (1989). Probabilistic semantics for nonmonotonic reasoning: a survey. In *KR '89*, pp. 505–516.

Shafer, G. (1976). *A Mathematical Theory of Evidence*. Princeton University Press.

Spohn, W. (1988). Ordinal conditional functions: a dynamic theory of epistemic states. In W. Harper and B. Skyrms (Eds.), *Causation in Decision, Belief Change and Statistics*, Volume 2, pp. 105–134. Reidel.

Wang, Z. and G. J. Klir (1992). *Fuzzy Measure Theory*. Plenum Press.

Weydert, E. (1994). General belief measures. In *UAI '94*, pp. 575–582.